\def\year{2020}\relax
\documentclass[letterpaper]{article} 
\usepackage{aaai20}  
\usepackage{times}  
\usepackage{helvet} 
\usepackage{courier}  
\usepackage[hyphens]{url}  
\usepackage{graphicx} 
\usepackage{graphicx}  
\usepackage{epsfig}
\usepackage{multirow}
\usepackage{subfigure}
\usepackage{amsmath}
\usepackage{amssymb}
\usepackage{array}
\usepackage{color}
\usepackage{verbatim}

\title{Monocular 3D Object Detection with Decoupled Structured Polygon Estimation and Height-Guided Depth Estimation}
\author{Yingjie Cai\textsuperscript{1}, 
\ Buyu Li\textsuperscript{1},
\ Zeyu Jiao\textsuperscript{2}, 
\ Hongsheng Li\textsuperscript{1},
\ Xingyu Zeng\textsuperscript{2},
\ Xiaogang Wang\textsuperscript{1} \\
\textsuperscript{1}The Chinese University of Hong Kong \quad
\textsuperscript{2}Sensetime Group Limited \qquad \\
{\tt\small caiyingjie@link.cuhk.edu.hk, \{jiaozeyu, zengxingyu\}@sensetime.com, \{byli, hsli, xgwang\}@ee.cuhk.edu.hk}
}
\begin{document}
\maketitle
\vspace{-0.5cm}
\begin{abstract}
Monocular 3D object detection task aims to predict the 3D bounding boxes of objects based on monocular RGB images. Since the location recovery in 3D space is quite difficult on account of absence of depth information, this paper proposes a novel unified framework which decomposes the detection problem into a structured polygon prediction task and a depth recovery task. Different from the widely studied 2D bounding boxes, the proposed novel structured polygon in the 2D image consists of several projected surfaces of the target object. Compared to the widely-used 3D bounding box proposals, it is shown to be a better representation for 3D detection. In order to inversely project the predicted 2D structured polygon to a cuboid in the 3D physical world, the following depth recovery task uses the object height prior to complete the inverse projection transformation with the given camera projection matrix. Moreover, a fine-grained 3D box refinement scheme is proposed to further rectify the 3D detection results. Experiments are conducted on the challenging KITTI benchmark, in which our method achieves state-of-the-art detection accuracy.
\end{abstract}
\vspace{-0.5cm}
\section{Introduction}\label{section:introduction}
3D object detection is an important computer vision task since it is an essential component of autonomous driving and robot perception to avoid collisions with surrounding objects.
Most existing 3D object detection methods heavily rely on LiDAR devices to obtain accurate and direct depth measurements. However, such sensors can not widely adopted due to the expensive cost and limited perception range ($\sim$100m). The farther away the objects are, the fewer and sparser depth measurements would be on the objects. In contrast, cameras are much cheaper and can be installed on any vehicles. This paper mainly focuses on 3D detection with monocular images.
\begin{figure}[t]
	\centering
	\includegraphics[width=1\linewidth]{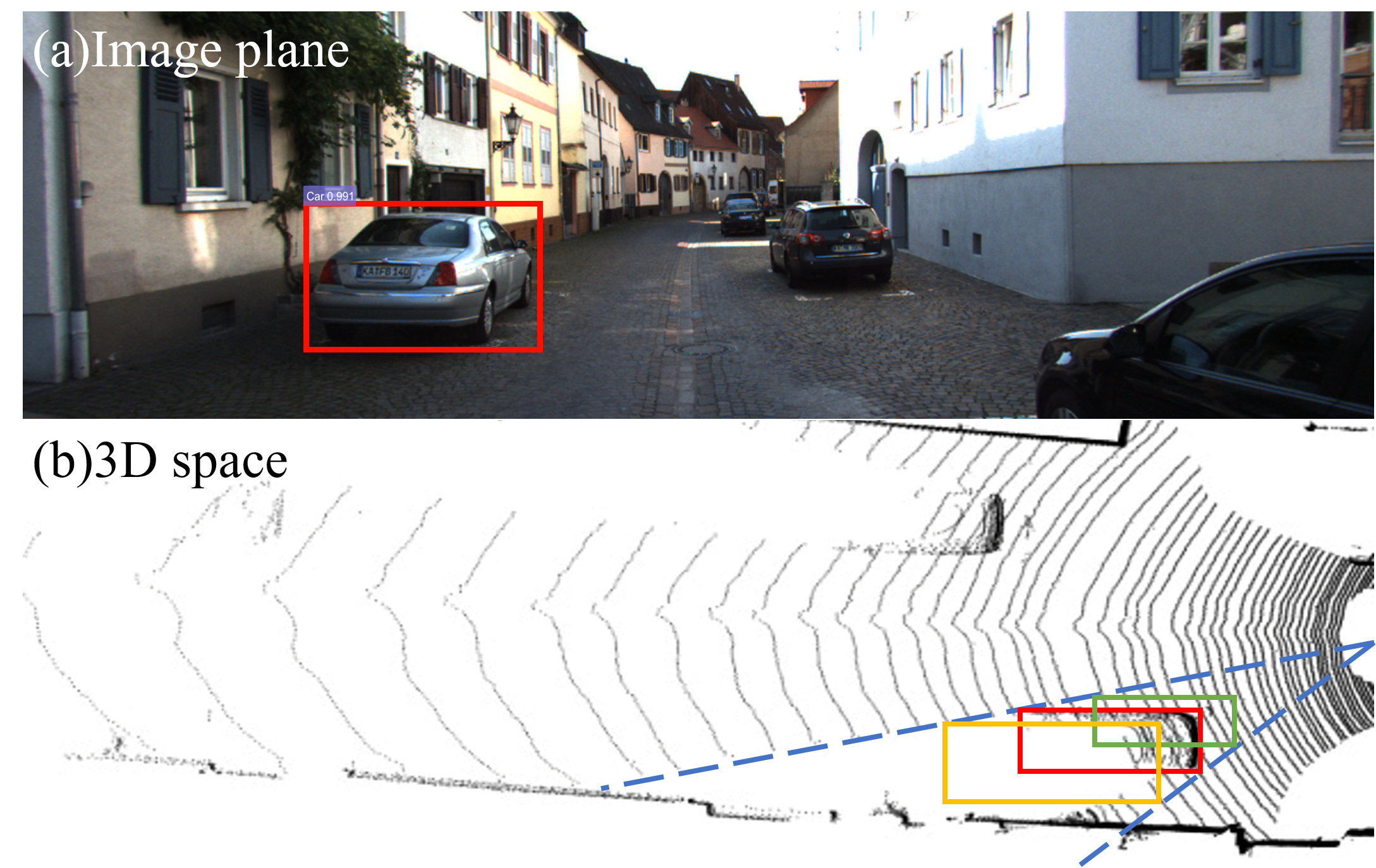}
	\caption{3D detection from 2D monocular images is challenging as even accurate 2D detection boxes (top) correspond to ambiguous 3D detection boxes (bottom). Best viewed in color.}
	\label{fig11}
\end{figure}
In general, a 3D bounding box can be described by 7 parameters in autonomous driving scenarios, i.e. the location $(x, y, z)$, size $(l, w, h)$ and orientation $\theta$ on the gr{}ound. For 3D detection from monocular images, recovering the location in 3D space is challenging on account of the absence of the accurate depth measurements. 
As illustrated in Fig. \ref{fig11}, given an accurate 2D bounding bo{}x of an object (Fig. \ref{fig11} (a)), its 3D location is still difficult to recover because one 2D box has an infinite number of corresponding 3D boxes (Fig. \ref{fig11} (b)) according to 2D-to-3D projection.
\begin{figure*}[h]
	\centering
	\includegraphics[width=1\linewidth]{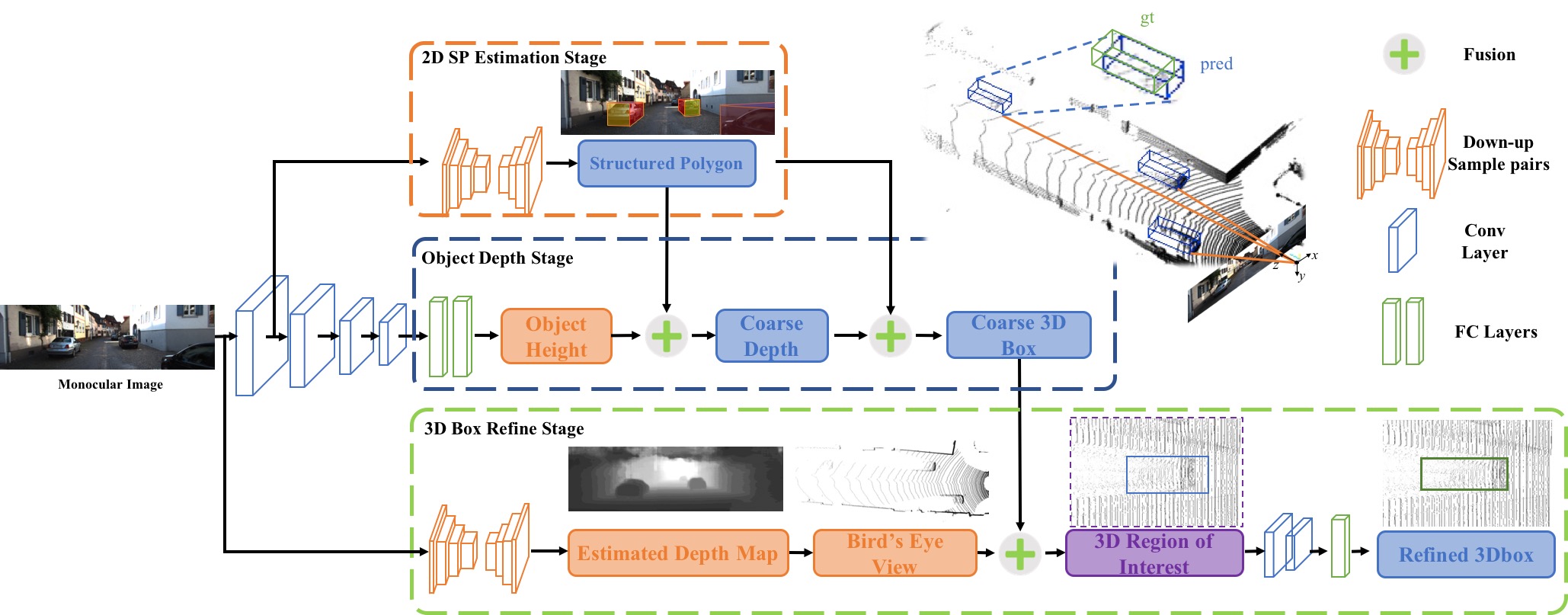}
	\caption{The overall framework \textbf{(\textit{Decoupled-3D})} decouples the monocular 3D object detection problem into sub-tasks. The overall network consists of three parts. (Top row) The 2D structured polygons are generated with a stacked hourglass network. (Middle row) Object depth stage utilizes 3D object height as a prior to recover missing depth of the object. (Bottom row) 3D box refine stage rectifies coarse 3D boxes using bird's eye view features in $3D$-$ROI$s. Best viewed in color.}
	\label{overview}
\end{figure*}
However, the projection of the 3D bounding box on the 2D image plane is unique and much easier to estimate with features in a 2D image. Since the projected 3D box follows prior knowledge of being a polygon consisting of several quadrilaterals (corresponding to the visible surfaces of the 3D box), we refer it as \textit{structured polygon} for convenience. The 3D box can be completely recovered given the structured polygon, depth and the projection matrix. As the camera projection matrix is generally known in auto-driving scenarios, the only additional information required is the depth of the object.

Inspired by the analysis above, we propose a novel framework that decomposes the 3D object detection task into a structured polygon prediction task and a depth estimation task. Different from the commonly used 2D bounding boxes in most previous works \cite{chen2016monocular}, \cite{mousavian20173d}, \cite{xu2018multi} and \cite{qi2018frustum}, the structured polygon can provide richer information for the 3D box recovery. Since the challenging depth estimation task is decoupled, the specific module can be designed to accurate tackle it. Moreover, the prediction of other parameters, e.g., the orientation of the object and the aspect ratio, would not be affected by depth estimation. Therefore, our framework has significant superiority over existing works that estimate all the 3D factors simultaneously \cite{chen20183d}, \cite{chen2017multi}, \cite{xu2018multi}, \cite{qi2018frustum}.

To obtain the structured polygon, we regress the projection points of the eight vertices of the cuboid via a stacked hourglass network. With the predicted structured polygon, we propose an efficient method to estimate the depth. Specifically, we use the \textit{object height} as a prior to compute the inverse projection transformation from the 2D image plane to 3D space with the given camera projection matrix. With the estimated depth and the obtained structured polygon, we can consequently recover the complete 3D box. 

The 3D box obtained from previous steps is actually a coarse estimation. We further propose a 3D box refinement scheme to rectify the coarse 3D boxes. We use the local features around and within the coarse 3D box for the refinement. The features are extracted from the bird's eye view map of the estimated depth map by a monocular depth estimation algorithm DORN \cite{FuCVPR18-DORN}.
In contrast to previous works that refine the 3D boxes with 2D image-level features such as MGR \cite{qin2019monogrnet} and MLF \cite{xu2018multi}, we refine the deviations of coarse boxes with bird's eye view map containing direct spatial information. The ingenious design can align the coarse boxes adaptively and substaintially enhances the accuracy.

To validate the effectiveness of our method, we perform thorough experiments on the challenging 3D object detection benchmark KITTI \cite{Geiger2012CVPR} and achieve new state-of-the-art performance under both $AP_{BEV}$ and $AP_{3D}$ metrics.
The contributions are summarized as following three-fold:
\begin{itemize}
\item A novel framework, which decomposes the challenging 3D detection problem into sub-tasks of image based structured polygon prediction and object depth estimation, is proposed. The two decomposed sub-tasks can be better tackled.
\item An efficient object depth estimation approach is proposed, which uses the \emph{object height} as a prior. Combined the depth with the structured polygon, coarse 3D boxes can be obtained.
\item A fine-grained 3D box refinement scheme is proposed. Different from the existing methods, we rectify the coarse boxes with bird's eye view map, which significantly improves the accuracy of the 3D boxes.
\end{itemize}
\section{Related work}
We briefly review recent works based on LiDAR data, stereo images and monocular images.\\
\textbf{LiDAR-based 3D Object Detection.} Most state-of-the-art 3D object detection methods reconstruct 3D bounding box using point clouds from LiDAR. \cite{luo2018fast} and \cite{zhou2018voxelnet} quantize the raw point cloud by using voxel grid and then feed the structured voxel grid to 2D or 3D CNN to detect 3D objects. \cite{qi2018frustum} and \cite{shi2019pointrcnn} directly exploit raw point cloud to generate 3D bounding boxes instead of quantizing to voxel grid with less information lossing. They respectively uses 2D bounding box and segmentation to lock effective point cloud and both encode point cloud via PointNet++ \cite{qi2017pointnet}. Our method focuses on monocular data setting and unavoidably suffers from the lack of accurate and direct depth measurements.\\
\textbf{Stereo-based 3D Object Detection.} There are several works are based on stereo vision. Stereo R-CNN \cite{li2019stereo} utilizes stereo RPN to detect 3D objects on left and right images simultaneously and tries novel pixel-level refinement based on stereo matching to refine 3D boxes. 3DOP \cite{chen20153d} assumes enormous 3D candidates and exploits ground-plane prior and object size to filter the candidates. Stereo images provide more information than monoculars. However, stereo setting has high requirements during the camera installation while our approach only needs a single image, which can be more flexible in real cases. \\
\textbf{Monocular-based 3D Object Detection.} More and more recent works are based on monocular images even through it is the most difficult. MGR \cite{qin2019monogrnet} and Mono3D \cite{chen2016monocular} encode RGB image feature to 2D CNN to regress 3D proposals and further refine the proposals with superimposed 2D features. Mono3D generates a diverse set of 3D candidate boxes first  and exploites ground plane prior and 2D cues including segmentation and object size to filter the candidates. Pseudo-LiDAR \cite{pseudo} transforms the depth map into pseudo point clouds and feeds the points into LiDAR based methods. \cite{kehl2017ssd}, \cite{pepik20153d} and \cite{chabot2017deep} adopt CAD models to build templates for better supervision meaning that more objects poses are prepared. Deep3Dbox \cite{mousavian20173d} leverages the geometry constrains between 3D and 2D bounding box to recover the 3D poses. These methods use more information from the superimposed 2D image level or additional CAD models to constrain. However, our method decomposes the problem in a novel approach and designs specific strategies for each subtasks to achieve better 3D object detection.
\section{Our Approach}
In this section, we present our proposed framework for 3D object detecting from monocular images. First, we introduce the overall formulation of our architecture. We then introduce of 3D coarse box estimation with structured polygon and depth estimation. Finally, we demonstrate a 3D box refinement stage to rectify coarse boxes. We name our method 3D object detection via decoupled tasks as \textit{Decoupled-3D}, as illustrated in Fig. \ref{overview}.

\subsection{Decoupled Tasks}\label{section:decouple}
As introduced, since the estimation of depth is the most strenuous part for monocular based 3D object detection, we decouple the depth from complicated 3D box estimation and decompose the task into structured polygon prediction and coarse-to-fine depth estimation sub-tasks. 
\begin{figure}[h]
	\centering
	\includegraphics[width=1\linewidth]{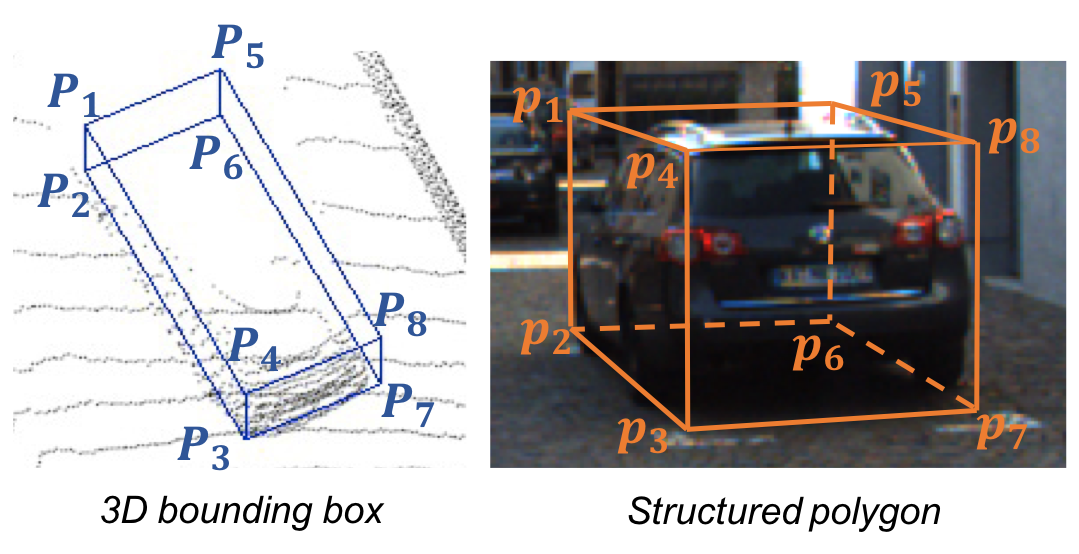}
	\caption{3D bounding box (left) and structured polygon (right). Best viewed in color.}
	\label{fig3}
\end{figure}

In 3D object detection, each object can be covered by a minimal cuboid in the 3D space, denoted by $B_{3d}$. A cuboid contains eight vertices $P_i=[X_i,Y_i,Z_i]^T$ $\in$ $\Re^3$, $i$=1,$\cdots$,8, as the left blue cuboid shown in Fig. \ref{fig3}. An object corresponds to a special \textit{structured polygon} on the 2D image plane via 3D-to-2D projection.
A structured polygon contains the eight vertices \{$p_i=[u_i,v_i]^T|i=1,\cdots,8$\} as the 2D vertices shown in Fig. \ref{fig3} (right). Given the camera intrinsic matrix $K$, the projection of a 3D vertex $P_i$ on the image plane is formulated as the following equation:
\begin{equation}
\label{transform}
K \cdot
\begin{bmatrix} X_i, Y_i, Z_i \end{bmatrix}^\mathrm{T}=
\begin{bmatrix} u_i, v_i, 1 \end{bmatrix}^\mathrm{T}\cdot Z_i ,
\end{equation}
where $Z_i$ is the depth of the vertex $P_i$. Given $p_i$ and $Z_i$, $P_i$ can be obtained as:
\begin{equation}
\label{decompose}
\begin{bmatrix} X_i, Y_i, Z_i \end{bmatrix}^\mathrm{T}=
K^{-1}\cdot
\begin{bmatrix} u_i, v_i, 1 \end{bmatrix}^\mathrm{T}\cdot Z_i .
\end{equation}
According to this equation, to estimate $B_{3D}$, we just need $K$, projected 2D vertices $p_i$ and the corresponding depth $Z_i$. As the camera intrinsic matrix $K$ is generally known, the remaining problem is the estimation of the 2D vertices of the structured polygon and their corresponding depths.
\subsection{2D Structured Polygon Estimation} \label{section:image3dbox}
In order to first obtain the locations of objects in the 2D image, we predict the 2D bounding box of each object by Faster RCNN \cite{ren2015faster}.

\begin{figure}[h]
	\centering
	\includegraphics[width=1\linewidth]{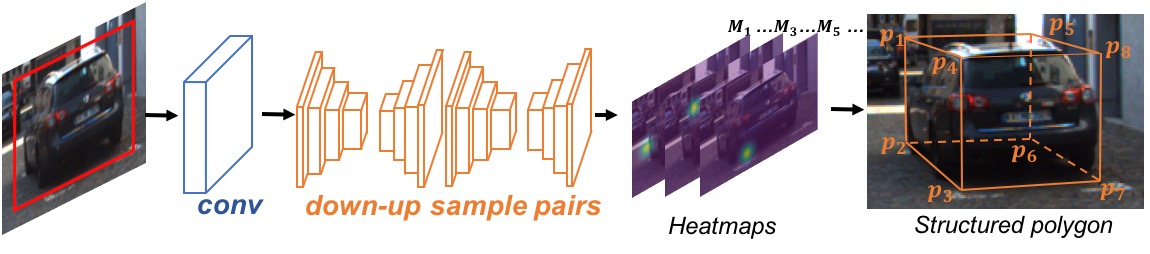}
	\caption{Structured polygon estimation aims to estimate the 2D locations of the projected vertices.}
	\label{fig4}
\end{figure}

For structured polygon estimation, we propose to regress the 2D image coordinates of 8 vertices based on the features extracted from the object area. However, it is still difficult to find accurate position in occlusion areas, texture-less regions and reflective surfaces. As shown in Fig. \ref{fig4}, the vertices are projected on texture-less background such as ground plane and wall without strong physical meaning. Solely applying the local feature is generally insufficient for accurate estimation in such challenging regions. Therefore, global context information should be incorporated to infer accurate positions of the vertices.

To capture more context information, we adopt a stacked hourglass architecture after several shallow convolution layers (res2 in ResNet). This architecture consists of two repeated top-down/bottom-up hourglass modules \cite{newell2016stacked}. The fusion of features from multi-scales can integrate both local and global features to obtain accurate vertices positions. Each of the eight projected vertices has a corresponding output heatmap from the network, as shown in Fig. \ref{fig4}. 
We use $M_i$ to denote the heatmap of the projected vertex $p_i$, and $M_i(u,v)$ is the value of the pixel $(u,v)$ on the heatmap, which represents the probability of the vertex locating at the this location.
The supervision of each vertex is a label map $\hat{M_i}$ with the ground truth position being one and others being zeros. We use the Euclidean loss function for training, which can help the model converge faster:
\begin{equation}
L_{sp} = \sum_i || M_i - \hat{M_i} ||_2^2
\end{equation}
During testing, the vertex position is estimated as the location with the highest probability.
\begin{equation}
\hat{p_i}=\mathop { \arg\max}_  {(u,v)} M_i(u,v)
\end{equation}
The label of eight projected vertices can be obtained via 3D-to-2D projection with Eq. (\ref{transform}) from 3D coordinates of vertices.
\subsection{Height-Guided Depth Estimation} \label{section:depth restoration based on projection rules}
The depth of an object is the most challenging parameter to estimate due to the fact that this information is missing after 3D-to-2D projection. Therefore, instead of directly regressing the strenuous depth from image-level features, we choose to recover the depth via camera projection principle. Based on the projection principle, we adopt a simple, but effective strategy for the missing depth via \textit{structured polygon} and a 3D physical prior, \textit{object height}.
\begin{figure}[h]
	\centering
	\includegraphics[width=1\linewidth]{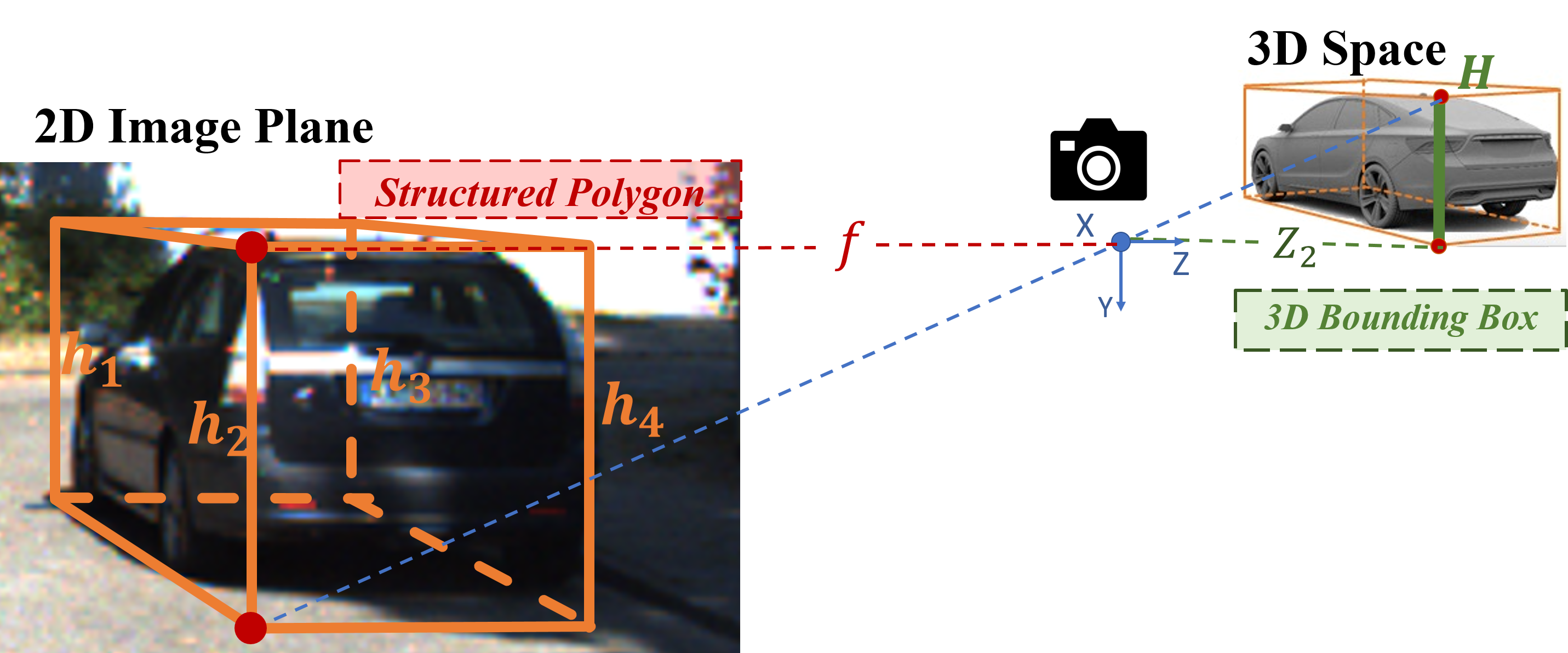}
	\caption{\textbf{Depth estimation based on object height prior.} Combine object height $H$ and corresponding pixel value $h$ to estimate object depth. Best viewed in color.}
	\label{depthcc}
\end{figure}

As shown in Fig. \ref{depthcc}, $f$ is the camera focal length, \textit{H} represents the 3D height of object and $h_j$ for $j$=1, 2, 3, 4 is the projected height of one vertical edge of the cuboid.
The height values of the four vertical edges in 3D space are the same, while the \textit{projected height} of the four vertical edges are different due to their different depths in the 3D space. Fig. \ref{depthcc} clearly shows the 3D-to-2D projection process of one vertical edge (i.e., $h_2$). Therefore, the corresponding depth ($Z_j$) of each vertical edge of the cuboid can be expressed as
\begin{equation}
\label{depth}
Z_j = f\cdot H/h_j \quad for \ j=1, 2, 3, 4
\end{equation}\\
where $h_j$ can be directly obtained from the estimated structured polygon, which is the pixel distance of two projected vertices.
For object height $H$, an intuitive idea is to use the average value $A_H$ obtained from the statistics of the height values in data set. However, the average height is not accurate enough for each instance. So we estimate the height of each object. Specifically, we pool the RoI feature of an object on the feature map from Res4, and then use 2 fully connected layers to predict the height. Instead of regressing the ground truth height $G_H$ directly, our regression target $t_H$ is the scale change:
\begin{equation}
t_H=log(G_H/A_H)
\end{equation}
The Smooth-L1 \cite{girshick2015fast} loss  function is adopted for the training of the regressor, since Smooth-L1 is less sensitive to outliers.
Further, according to Eq. (\ref{transform}), the coordinates of eight vertices in 3D space can be obtained via generated projected vertices and depths.

With the eight 3D vertices, we use an average operation to obtain a coarse 3D box. Specifically, in KITTI dataset the location of an object is defined as the position of the bottom center of its 3D bounding box, so we use the average of the midpoints of the diagnal $P_2P_7$ and $P_3P_6$ to estimate the location $(x, y, z)$. The $l$ can be calculated using the average of distances of $P_2P_3$, $P_6P_7$, $P_1P_4$ and $P_5P_8$. $h$ and $w$ are calculated using the similar way. Orientation $\theta$ comes from the average of four vertors $\overrightarrow{P_3P_2}$, $\overrightarrow{P_7P_6}$, $\overrightarrow{P_4P_1}$ and $\overrightarrow{P_8P_5}$.
\subsection{3D Box Refinement} \label{section:Mono 3D Box refinement}
\begin{figure}[h]
	\centering
	\includegraphics[width=1\linewidth]{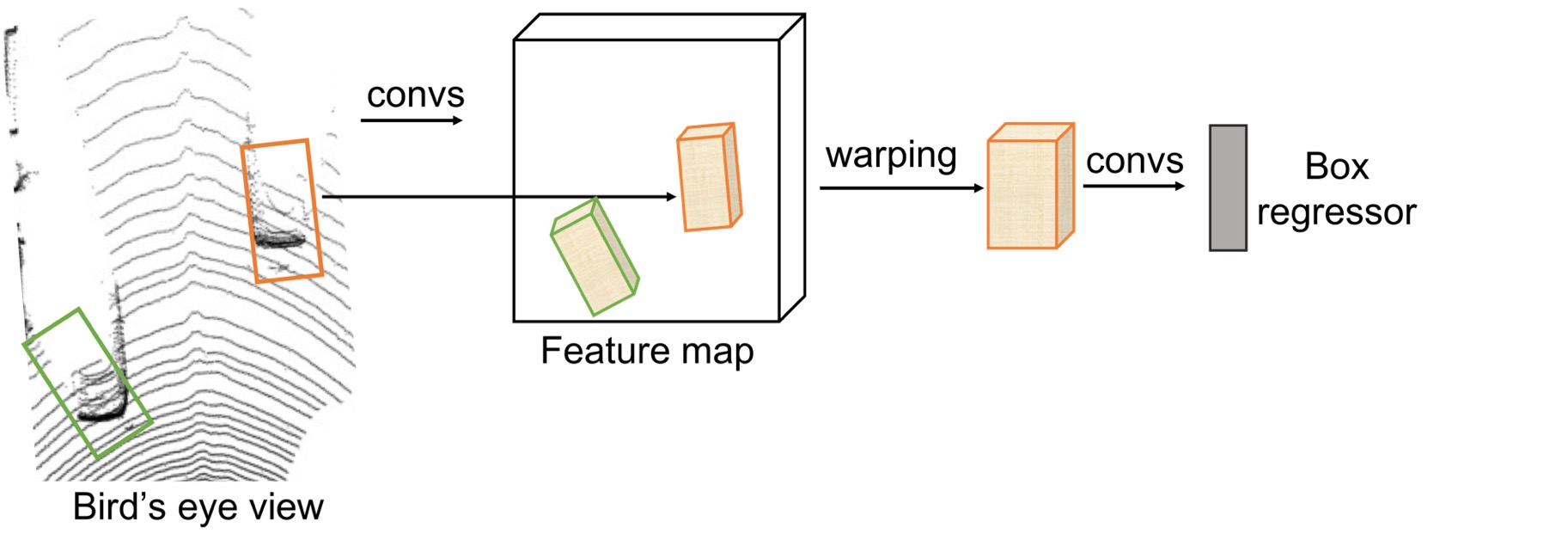}
	\caption{\textbf{3D box Refinement.} Rectify coarse boxes with bird's eye view map.}
	\label{ama}
\end{figure}

The 3D box obtained from previous steps is actually a coarse estimation. But the error is usually minor and the ground-truth is just located nearby. 
As shown in bird’s eye view of Fig. \ref{ama} (left), the predictions have deviations about 1$m$.

Based on this fact, the proposed method tailors a fine-grained refinement scheme for 3D detection. Different from existing methods exploiting image-level features or front-view depth map, we leverage bird's eye view map, which contains direct spatial information, to rectify coarse boxes. The well-designed scheme can adjust coarse 3D boxes to better locations in nearby region. For convenience, we refer this Fine-Grained refinement strategy as \textit{FG}.

Bird's eye view maps are transformed from predicted monocular depth maps using DORN \cite{FuCVPR18-DORN} . The details for the transformation process are outlined in section. \ref{section:Implementation}. 
We take an entire BEV map and a set of coarse boxes as input. The whole map is processed by a CNN to produce a feature map. Then, for each object we adopt an warping operation of the region on the feature map to extract a fixed-length feature vector. The region has 2x size of the coarse box to ensure that targets are within this region. Since this region contains direct 3D information, we named it $3D$-$ROI$ for convenience. Each feature vector is fed into a sequence of convolution layers followed by a fully-connected layer that outputs residual values $(\delta x,\delta y,\delta z,\delta l,\delta w,\delta h,\delta \theta)$ based on the coarse 3D box. A Smooth-L1 loss is used for the training of this network.

In this way, the deviations of coarse boxes are removed, which surges the performance. We argue that bird's eye view map with direct 3D spatial information is much more suitable for fine-grained refinement in 3D task than 2D image-level features or front-view depth map. Detailed analysis and comparisons are in section. \ref{section:Ablation Study}.
\section{Experiments}
We evaluate the proposed method on the KITTI object detection benchmark \cite{Geiger2012CVPR} and split training images into \emph{training set} and \emph{validation set} following the commnly used train/val split mentioned in 3DOP \cite{chen20183d}. In the KITTI dataset, objects are divided into three difficulty regimes: \emph{easy}, \emph{moderate} and \emph{hard}, according to the 2D box height, occlusion and truncation degrees following the KITTI official standard.
For all experiments, we follow most previous methods to focus on vehicle category as it has the majority of samples in the dataset.
\subsection{Implementation Details} \label{section:Implementation}
ResNet-50 \cite{He2015} is selected as our basic backbone to extract features, which is initialized by pre-trained weights on ImageNet \cite{russakovsky2015imagenet}. The initial learning rate is 0.001 for the previous 30K interactions and then 0.0001 for another 10K iterations. For the 3D box refinement, we keep depth points in the following range of the bird's eye view:
$$ -25m\leq X \leq 25m, -1.5m\leq Y\leq 4.09m, 0\leq Z\leq 50m$$
The predicted depth map is from a monocular based method DORN \cite{FuCVPR18-DORN}. The row and column represent left-right  (i.e., $X$) direction and the depth (i.e., $Z$) direction respectively, and the top-down value, $Y$ is 
mapped to a slice at each $(X, Y)$ location on bird’s eye view maps.
The width and height of $3D$-$ROI$ are set as 256 and 456 respectively. The size is computed from the statistics to ensure targets are within this region. 
For the object height, we use 1.46\emph{m} as the average value.
\subsection{Comparison with Other Methods} \label{section:3D Object Detection on KITTI}
\begin{table*}
\centering
\caption{\textbf{Bird's eye view localization and 3D detection performance:} Average Precision in bird's eye view ($AP_{BEV}$) and Average Precision of 3D boxes ($AP_{3D}$) on KITTI $validation$ set.}
\label{core}
\begin{tabular}{|c||c|c|c|c|c|c|c|c|c|c|c|c|}
\hline
\multirow{2}{*}{Method}
&\multicolumn{3}{|c|}{IoU=0.5 $AP_{BEV}$} & \multicolumn{3}{|c|}{IoU=0.5 $AP_{3D}$} &\multicolumn{3}{|c|}{IoU=0.7 $AP_{BEV}$} & \multicolumn{3}{|c|}{IoU=0.7 $AP_{3D}$}\\
\cline{2-13}
& Easy & Mod. & Hard &Easy & Mod. & Hard &Easy & Mod. & Hard &Easy & Mod. & Hard \\
\hline
Mono3D& 30.50 & 22.93 & 19.16 & 25.19 & 18.20 & 15.52 & 5.22 & 5.19 & 4.13 & 2.53 & 2.31 & 2.31\\
Deep3Dbox &31.12 &22.53&18.12& 26.15&19.42&14.62 & 9.33 & 6.71 & 5.11 & 5.49 & 3.96 & 2.92\\
MLF &55.02 &36.73&31.27& 47.88&29.48&26.42&22.03&13.63&11.60&10.53&5.69&5.39\\
ROI-10D&-&-&-&-&-&-&14.76&9.55&7.57&10.25&6.39&6.18\\
GS3D&-&-&-&33.11&27.16&23.57&-&-&-&11.63&10.51&10.51\\
MGR &53.29 &37.55 &	30.46 &50.51&36.97&30.82&21.67& 14.93 & 12.34&13.88&10.19&7.62\\
MonoPSR&56.97&43.39&36.00&49.65&41.71&29.95&20.63&18.67&14.45&12.75&11.48&8.59\\
Pseudo-LiDAR&70.8&49.4&42.7&66.3&42.3&38.5&40.6&26.3&22.9&\textbf{28.2}&18.5&\textbf{16.4}\\
\hline
Ours & \textbf{73.22} & \textbf{54.31} & \textbf{45.97} & \textbf{69.40} & \textbf{50.50} & \textbf{42.46} & \textbf{44.42} & \textbf{29.69} &\textbf{24.60} & 26.95 & \textbf{18.68} & 15.82 \\
\hline
\end{tabular}
\end{table*}

\begin{table}[h]
\centering
\caption{Average precision for bird's eye view localization and 3D detection on KITTI \emph{test} set.}
\label{testing}
\begin{tabular}{|c| |p{0.65cm}<{\centering} |p{0.65cm}<{\centering} |p{0.65cm}<{\centering} |p{0.65cm}<{\centering} |p{0.65cm}<{\centering} |p{0.65cm}<{\centering}|}
\hline
\multirow{2}{*}{Method}
&\multicolumn{3}{|c|}{IoU=0.7 $AP_{BEV}$} & \multicolumn{3}{|c|}{IoU=0.7 $AP_{3D}$} \\
\cline{2-7}
& Easy & Mod. & Hard &Easy & Mod. & Hard  \\
\hline
ROI-10D &9.78& 4.91& 3.74& 2.02& 4.32& 1.46\\
GS3D&8.41 &6.08 &4.94 &2.90 &4.47 &2.47\\
MGR&18.19 &11.17& 8.73& 9.61& 5.74& 4.25\\
MonoPSR&18.33 &12.58 &9.91 &10.76 &7.25 &\textbf{5.85}\\
\hline
Ours&
\textbf{24.62} &\textbf{14.66} &\textbf{11.46} &\textbf{11.68} &\textbf{7.28} &5.69\\
\hline
\end{tabular}
\end{table}
We compare with state-of-the-art monocular based methods including Mono3D \cite{chen2016monocular}, Deep3Dbox \cite{mousavian20173d}, MLF \cite{xu2018multi}, ROI-10D \cite{roi}, GS3D \cite{li2019gs3d}, MGR \cite{qin2019monogrnet}, MonoPSR \cite{ku2019monocular} and Pseudo-LiDAR \cite{pseudo}.\\
\\
\textbf{Metrics.} The proposed method is evaluated by Average Precision on both bird's eye view ($AP_{BEV}$) and 3D detection ($AP_{3D}$) metrics. $AP_{BEV}$ evaluates whether the location of prediction is accurate by calculating the intersection over union (IoU) with ground-truth boxes from bird's eye view. This performance of this metric is critical for autonomous driving to avoid collision. $AP_{3D}$ counts the intersection of two cuboids (i.e., predicted box and ground truth) and adds object height and up-down information based on location.\\
\\
\textbf{Bird's Eye View Evaluation.} $AP_{BEV}$ evaluates the projection of the 3D box on bird's eye view. For comprehensive comparison, we experiment with two IoU thresholds (0.5 and 0.7) following exsting methods. Just as shown in Tab. \ref{core}, our method outperforms state-of-the-art monocular based methods at at both IoUs and surpasses Pseudo-LiDAR \cite{pseudo} by 4.91\% for 0.5 IoU and 3.39\% for 0.7 IoU under \emph{moderate} level respectively.\\
\\
\textbf{3D Detection Evaluation.} For $AP_{3D}$, we also perform evaluations under the two IoUs. Compared to $AP_{BEV}$, $AP_{3D}$ expands from bird's eye view plane to 3D space and calculates the intersection over union with 3D ground-truth boxes. As show in Tab. \ref{core}, our method outperforms state-of-the-art monocular methods in all difficulties for 0.5 IoU and surpasses Pseudo-LiDAR \cite{pseudo} by 8.20\% for \emph{moderate} level. The results of 0.7 IoU have certain gap with Pseudo-LiDAR.\\
\\
\textbf{Results on Test Set.} We submit our results to the KITTI test server for evaluation and compare with all monocular based published methods on the test set. 
As shown in Tab. \ref{testing}, the results show that our method outperforms the previous methods by significant margins in terms of all metrics, which proves the effectiveness of our method. Compared to the latest state-of-the-art method, our $AP_{BEV}$ has an average improvement of 3.31\%, and $AP_{3D}$ increases by 1\% at $easy$ level.
\subsection{Ablation Study} \label{section:Ablation Study}
In this section, we conduct ablation experiments to validate the effectiveness of different components of our overall framework. All comparison are engaged on \emph{validation set}.
\\
\textbf{Benefits of Decoupled Tasks.}
As introduced, we decompose the 3D box estimation problem into sub-tasks. To better evaluate the contribution of decoupled tasks, we compare our coarse and final results with regressing all variables simultaneously. As shown in Tab. \ref{ablation-archi}, the performance of jointly regressing all parameters is far worse than our decoupled strategy. Even compared with the coarse results, there is a drop of 10\% in terms of $AP_{BEV}$ and $AP_{3D}$.
\begin{table}[h]
\centering
\caption{\textbf{Ablation study of decoupled tasks.} Jointly represents all variables regressed simultaneously.}
\label{ablation-archi}
\begin{tabular}{|c||p{0.6cm}|p{0.6cm}|p{0.6cm}|p{0.6cm}|p{0.6cm}|p{0.6cm}|}
\hline
\multirow{2}{*}{Method}
&\multicolumn{3}{|c|}{IoU=0.5 $AP_{BEV}$} & \multicolumn{3}{|c|}{IoU=0.5 $AP_{3D}$}\\
\cline{2-7}
& Easy & Mod. & Hard &Easy & Mod. & Hard \\
\hline
Jointly & 16.44&12.05&9.90&6.84&4.50&4.13\\
\hline
Coarse & 26.42&	20.91&	17.93&	19.67&	16.36&	13.87\\
\hline
Ours & 
\textbf{73.22} & \textbf{54.31} & \textbf{45.97} & \textbf{69.40} & \textbf{50.50} & \textbf{42.46}\\
\hline
\end{tabular}
\end{table}
\\
\textbf{Benefits of Structured Polygon.}
We add an experiment of utilizing bird’s eye view (BEV) features to regress coarse 3D boxes instead of using structured polygon (SP). As shown in Tab. \ref{ablation-sp}, the model with structured polygon outperforms the one with BEV by 15.70\% and 22.62\% in terms of $AP_{BEV}$ and $AP_{3D}$ at \textit{moderate} level, which demonstrates the contribution of structured polygon.\\
\textbf{Benefits of Height-Guided Depth Estimation.}
As introduced in section. \ref{section:depth restoration based on projection rules}, we propose a simple, but effective height-guided depth recovery strategy. To verify the effectiveness, we regress the depth directly and use mean depth error to compare the two strategies. As shown in Tab. \ref{ablation-depth}, the mean depth error of height-guided depth recovery strategy is 1.21\textit{m}, while directly regressing depth is 2.41\textit{m} almost twice as much as the former. The depth inferred from height outperforms regressing directly, which is due to that the height-guided strategy tackles the problem by utilizing the stable physical prior \textit{object height}.
\begin{table}[h]
\centering
\caption{\textbf{Ablation study of structured polygon.} BEV represents coarse 3D boxes regressed with bird’s eye view map.}
\label{ablation-sp}
\begin{tabular}{|c||p{0.6cm}|p{0.6cm}|p{0.6cm}|p{0.6cm}|p{0.6cm}|p{0.6cm}|}
\hline
\multirow{2}{*}{Method}
&\multicolumn{3}{|c|}{IoU=0.5 $AP_{BEV}$} & \multicolumn{3}{|c|}{IoU=0.5 $AP_{3D}$}\\
\cline{2-7}
& Easy & Mod. & Hard &Easy & Mod. & Hard \\
\hline
BEV & 56.19  &38.61 &31.94 &40.03& 27.88 &22.70\\
\hline
SP &\textbf{73.22}& \textbf{54.31}& \textbf{45.97} &\textbf{69.40} &\textbf{50.50} &\textbf{42.46} \\
\hline
\end{tabular}
\end{table}

\begin{table}[h]
\centering
\caption{\textbf{Ablation study of height-guided depth estimation.} The lower the mean depth error value, the better the results.}
\label{ablation-depth}
\begin{tabular}{|c|c|c}
\hline
Method & mean depth error \\
\hline
regress directly & 2.41\textit{m} \\
height-guided & \textbf{1.21}\textit{m} \\
\hline
\end{tabular}
\end{table}
\noindent
\textbf{Benefits of 3D box refinement.}
As mentioned in section. \ref{section:Mono 3D Box refinement}, we propose a tailored refinement scheme for 3D task and argue that bird's eye view map with 3D spatial information is much more suitable than 2D image-level features and front-view depth map for fine-grained refinement in the 3D space. For comprehensive comparison, we refine coarse boxes with 2D image features and front-view depth map respectively. As shown in Tab. \ref{ablation-bird}, the impacts of 2D image-level features (i.e., $+img$) and front-view depth map (i.e., $+fv$) are basically the same. The results of 3D detection are all better than our coarse results and $AP_{3D}$ improves about 1.3\% at \textit{hard} level. The performance of localization is basically unchanged. While with the help of fine-grained 3D box refinement (i.e.,$FG$), $AP_{BEV}$ and $AP_{3D}$ have been substantially improved, which proves the effectiveness of $FG$ successfully capturing bird's eye view map.
\begin{table}[h]
\centering
\caption{\textbf{Ablation study of 3D box refinement.} Comparison of different refinement strategies. ($+img$)  and ($+fv$) represent refining the coarse results via image-level features and front-view depth map respectively.}
\label{ablation-bird}
\begin{tabular}{|c||p{0.6cm}|p{0.6cm}|p{0.6cm}|p{0.6cm}|p{0.6cm}|p{0.6cm}|}
\hline
\multirow{2}{*}{Method}
&\multicolumn{3}{|c|}{IoU=0.5 $AP_{BEV}$} & \multicolumn{3}{|c|}{IoU=0.5 $AP_{3D}$}\\
\cline{2-7}
& Easy & Mod. & Hard &Easy & Mod. & Hard \\
\hline
Coarse & 26.42&	20.91&	17.93&	19.67&	16.36&	13.87\\
\hline
$+img$&26.36&	20.74&	17.83&	21.82&	16.95&	15.17\\
$+fv$&26.32&20.70&17.80&21.81& 16.93&15.14\\
$+FG$ 
& \textbf{73.22} & \textbf{54.31} & \textbf{45.97} & \textbf{69.40} & \textbf{50.50} & \textbf{42.46}\\
\hline
\end{tabular}
\end{table}
\begin{figure*}[t]
	\centering
	\includegraphics[width=0.95\linewidth]{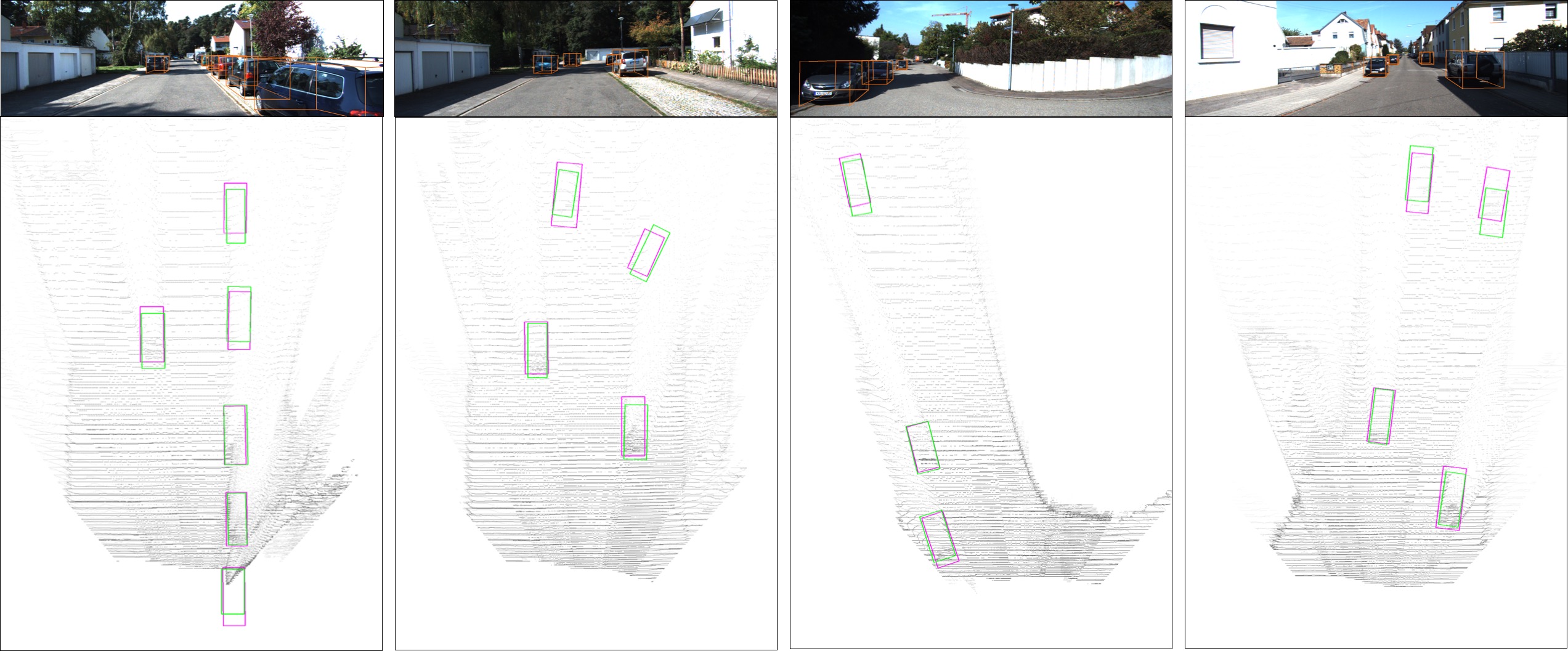}
	\caption{\textbf{Qualitative Results.} Top: structured polygons. Bottom: 3D bounding boxes in bird's eye view. Camera center is located at bottom center. Predicted 3D bounding boxes are drawn in pink, while ground truths are in green. Best viewed in color.}
	\label{resultsshow}
\end{figure*}

We also conduct an additional experiment to verify the contribution of $FG$. As shown Tab. \ref{ablation-other}, when applied on other methods like Mono3D \cite{chen2016monocular} and MGR \cite{qin2019monogrnet}, $FG$ can also improve them with significant margins.
\begin{table}[h]
\centering
\caption{\textbf{Ablation study of 3D box refinement.} Results for other methods refined similarly.}
\label{ablation-other}
\begin{tabular}{|c||p{0.6cm}|p{0.6cm}|p{0.6cm}|p{0.6cm}|p{0.6cm}|p{0.6cm}|}
\hline
\multirow{2}{*}{Method}
&\multicolumn{3}{|c|}{IoU=0.5 $AP_{BEV}$} & \multicolumn{3}{|c|}{IoU=0.5 $AP_{3D}$}\\
\cline{2-7}
& Easy & Mod. & Hard &Easy & Mod. & Hard \\
\hline
Mono3D & 30.50 & 22.93 & 19.16 & 25.19 & 18.20 & 15.52\\
Mono3D+$FG$& 54.12 & 37.67 & 32.20 & 41.04 &	29.05 &	24.59\\
\hline
MGR & 53.29 &37.55 & 30.46 & 50.51 & 36.97 & 30.82 \\
MGR+$FG$ & 65.87 &	49.01 & 40.93 & 60.95 & 43.80 & 36.27 \\
\hline
\end{tabular}
\end{table}

The remarkable margins for our coarse to final results is due to following two reasons.
One is about the elaborately designed refinement scheme, which leverages bird's eye view map. Compared to other formats, our design with direct spatial information is more suitable for 3D detection.

\begin{figure}[h]
	\centering
	\includegraphics[width=0.85\linewidth]{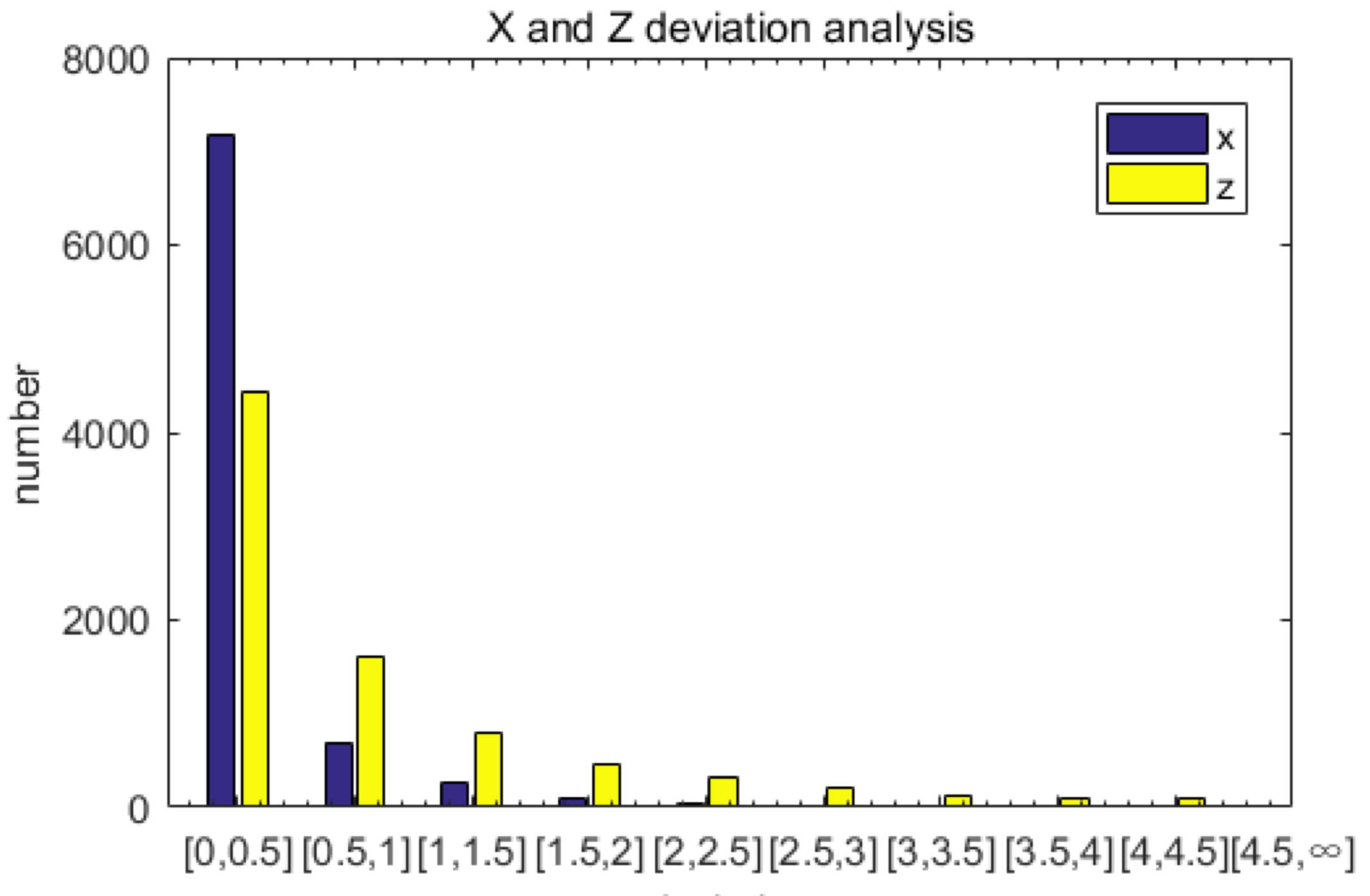}
	\caption{\textbf{Deviation analysis of $X$ and $Z$}. The figure shows the number of absolute deviation of $X$ and $Z$ of the coarse boxes compared with ground-truth boxes in different interval segments. Best viewed in color.}
	\label{XZ}
\end{figure}

Another is that most coarse boxes are actually not far from the corresponding targets. As shown in Fig. \ref{XZ}, most deviations of coarse $X$ and $Z$ are within 1$m$. As long as coarse boxes are shifted suitably, the accuracy can be increased substantially.
\subsection{Qualitative Results} \label{section:Qualitative Results}
We show some quantitative results in Fig. \ref{resultsshow}, where structured polygons are visualized in top row and the 3D boxes are showed in bottom on bird's eye view. The pink and green boxes represent ground-truth boxes and predicted results respectively. As seen from the visualization results, our method can accurately predict the boxes in different locations and orientations only based on monocular images.

\section{Conclusion}
In this paper, we propose an efficient monocular 3D object detection framework which decomposes the complicated 3D detection problem into a structured polygon prediction task and a following depth recovery task. The former task uses a stacked top-down/bottom-up hourglass network to build the structured polygon with a pretty high precision. The following depth recovery task utilizes the object height prior to inversely project the structured polygon to a cuboid in the 3D space. Moreover, a fine-grained refinement scheme is then adopted to rectify the deviations, which uses the local feature from the bird's eye view transformed from the prediction of a monocular depth estimation algorithm. Experiments on the KITTI benchmark proves the effectiveness of our proposed framework.
\section{Acknowledgment}
This work is supported in part by SenseTime Group Limited, in part by the General Research Fund through the Research Grants Council of Hong Kong under Grants CUHK14202217, CUHK14203118, CUHK14207319 and CUHK14208619.
\clearpage
{\small
\bibliographystyle{aaai}
\bibliography{egbib}
}
\end{document}